\documentclass[10pt,twocolumn,letterpaper]{article}

\usepackage{iccv}
\usepackage{times}
\usepackage{epsfig}
\usepackage{graphicx}
\usepackage{amsmath}
\usepackage{amssymb}
\usepackage{subfigure}
\usepackage{diagbox}
\usepackage{overpic}
\usepackage{adjustbox}
\usepackage{breqn}
\usepackage{multirow}
\usepackage{multicol}
\usepackage{setspace}

\usepackage{enumitem}

\usepackage[pagebackref=true,breaklinks=true,letterpaper=true,colorlinks,bookmarks=false]{hyperref}

\iccvfinalcopy 
\def\iccvPaperID{0000} 

\ificcvfinal\fi
\begin{document}

\title{AC-Lite : A Lightweight Image Captioning Model for Low-Resource Assamese Language}

\author{
Pankaj Choudhury \\
Indian Institute of Technology Guwahati\\
Assam India\\
{\tt\small pankajchoudhury@iitg.ac.in}
\and
Yogesh Aggarwal\\
Indian Institute of Technology Guwahati\\
Assam India\\
{\tt\small yogesh\_aggarwal@iitg.ac.in}
\and
Prabhanjan Jadhav\\
Indian Institute of Technology Guwahati\\
Assam India\\
{\tt\small j.prabhanjan@iitg.ac.in}
\and
Prithwijit Guha \\
Indian Institute of Technology Guwahati\\
Assam India\\
{\tt\small pguha@iitg.ac.in}
\and
Sukumar Nandi \\
Indian Institute of Technology Guwahati\\
Assam India\\
{\tt\small sukumar@iitg.ac.in}
}

\maketitle

\begin{abstract}
Most existing works in image caption synthesis use computation heavy deep neural networks and generates image descriptions in English language. This often restricts this important assistive tool for widespread use across language and accessibility barriers. This work presents AC-Lite, a computationally efficient model for image captioning in low-resource Assamese language. AC-Lite reduces computational requirements by replacing computation-heavy deep network components with lightweight alternatives. The AC-Lite model is designed through extensive ablation experiments with different image feature extractor networks and language decoders. A combination of ShuffleNetv2x1.5 with GRU based language decoder along with bilinear attention is found to provide the best performance with minimum compute. AC-Lite was observed to achieve an 82.3 CIDEr score on the COCO-AC dataset with 2.45 GFLOPs and 22.87M parameters.

\end{abstract}

\section{Introduction}
\label{sec:intro}
Deep neural networks have elevated the capability of machines for image understanding and natural language modeling. Automatic Image Captioning (AIC) is a research area that enables machines to automatically generate meaningful textual descriptions for input images. This capability is essential for multimodal AI applications, including assistive technologies for visually impaired individuals, intelligent human-machine interfaces, and visual content retrieval~\cite{bai2018survey}.  However, despite significant advancements in image captioning research ~\cite{xu2015show,vinyals2015show,anderson2018bottom,huang2019attention,herdade2019image}, the practical deployment of the state-of-art models remain a major challenge. This can be attributed to the prioritization of achieving arbitrarily high overall performance at the cost of computational efficiency. This often renders most models unsuitable for certain real-world applications that require on-device processing and low-latency inference on energy-efficient edge devices.

\begin{figure}
\centering
\includegraphics[width=\columnwidth, height=1.3cm]{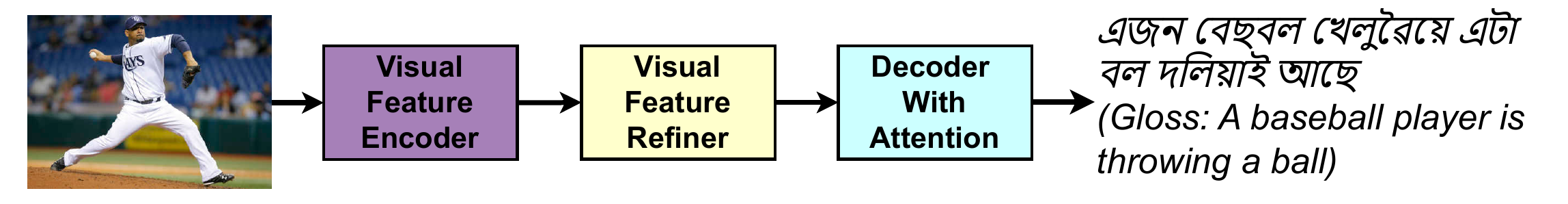}
\caption{\small{Various components of an Image Captioning System.}}
\label{fig:ic-components}
\end{figure}

Early image captioning models have used template-based and retrieval-based approaches to generate captions. These early approaches have either filled blank spaces in a fixed sentence template or retrieved similar captions from a prefix caption database. Though these approaches were simple, the generated captions were of fixed size, or the captions were often irrelevant to the image content. The introduction of deep learning-based image captioning models could mitigate these issues up to some extent~\cite{hossain2019comprehensive}. The deep learning-based models mainly consist of three main components -- (a) \emph{Visual Feature Encoder}, (b) \emph{Visual Feature Refiner}, and (c) \emph{Decoder With Attention} (see \autoref{fig:ic-components}) \cite{hossain2019comprehensive}. The deep learning based models~\cite{xu2015show,vinyals2015show,anderson2018bottom,huang2019attention,herdade2019image} have achieved impressive performance by leveraging computation-heavy visual feature extractors. However, these advancements came with a significant increase in computational demands and memory footprint. \autoref{tab:model-computation} highlights notable deep learning-based image captioning models along with computations (measured in Billion FLOPs or GFLOPs) and parameter counts (in Millions)\footnote{FLOPs and Parameters are computed using fvcore library (https://github.com/facebookresearch/fvcore).}. As shown in \autoref{tab:model-computation}, models like \emph{Show and Tell}~\cite{vinyals2015show} and \emph{Show Attend and Tell}~\cite{xu2015show} used simpler architectures and achieved comparable performance. However, models like \emph{Show Attend and Tell} used a heavy Convolutional Neural Network (CNN) such as VGG19~\cite{vgg16} as the visual feature encoder, which increased the overall computation. Additionally, models like \emph{BUTD}~\cite{anderson2018bottom} and \emph{AoANet}~\cite{huang2019attention} relied on FasterRCNN~\cite{ren2015faster} as the visual feature encoder. The computation-heavy FasterRCNN contributed significantly to the overall computation. 

\begin{table*}[]
\centering
\caption{Component-wise computational complexity (in Giga FLOPs) and parameter (in Millions) comparison of various deep learning-based Image Captioning models. Note that the FLOPs and Parameter calculations are based on the hyper-parameters mentioned in the respective papers.}
\label{tab:model-computation}
\begin{adjustbox}{width=\textwidth}{
\begin{tabular}{|c|ccc|ccc|ccc|cc|}
\hline
\multirow{2}{*}{\textbf{\begin{tabular}[c]{@{}c@{}}Model \\ Categories\end{tabular}}} & \multicolumn{3}{c|}{\textbf{Visual Feature Extractor}} & \multicolumn{3}{c|}{\textbf{Visual Feature Refiner}} & \multicolumn{3}{c|}{\textbf{Decoder with Attention}} & \multicolumn{2}{c|}{\textbf{Total}} \\ \cline{2-12} 
 & \multicolumn{1}{c|}{\textbf{Model}} & \multicolumn{1}{c|}{\textbf{GFLOPs}} & \textbf{Params (M)} & \multicolumn{1}{c|}{\textbf{Model}} & \multicolumn{1}{c|}{\textbf{GFLOPs}} & \textbf{Params (M)} & \multicolumn{1}{c|}{\textbf{Model}} & \multicolumn{1}{c|}{\textbf{GFLOPs}} & \textbf{Params (M)} & \multicolumn{1}{c|}{\textbf{GFLOPs}} & \textbf{Params (M)} \\ \hline
Show and Tell & \multicolumn{1}{c|}{GoogLeNet} & \multicolumn{1}{c|}{1.5} & 6.6 & \multicolumn{1}{c|}{-} & \multicolumn{1}{c|}{-} & - & \multicolumn{1}{c|}{LSTM} & \multicolumn{1}{c|}{0.131} & 13.4 & \multicolumn{1}{c|}{1.631} & 20 \\ \hline
\begin{tabular}[c]{@{}c@{}}Show, Attend\\  and Tell\end{tabular} & \multicolumn{1}{c|}{VGG19} & \multicolumn{1}{c|}{19.5} & 20 & \multicolumn{1}{c|}{-} & \multicolumn{1}{c|}{-} & - & \multicolumn{1}{c|}{LSTM} & \multicolumn{1}{c|}{0.401} & 14.45 & \multicolumn{1}{c|}{19.901} & 34.45 \\ \hline
BUTD & \multicolumn{1}{c|}{FasterRCNN} & \multicolumn{1}{c|}{117.33} & 63.63 & \multicolumn{1}{c|}{-} & \multicolumn{1}{c|}{-} & - & \multicolumn{1}{c|}{UpDown} & \multicolumn{1}{c|}{0.945} & 52.12 & \multicolumn{1}{c|}{118.275} & 115.75 \\ \hline
AoANet & \multicolumn{1}{c|}{FasterRCNN} & \multicolumn{1}{c|}{117.33} & 63.63 & \multicolumn{1}{c|}{AoA refiner} & \multicolumn{1}{c|}{4.527} & 44.09 & \multicolumn{1}{c|}{\begin{tabular}[c]{@{}c@{}}AoA \\ Decoder\end{tabular}} & \multicolumn{1}{c|}{0.586} & 41.48 & \multicolumn{1}{c|}{122.443} & 149.2 \\ \hline
ORT & \multicolumn{1}{c|}{FasterRCNN} & \multicolumn{1}{c|}{117.33} & 63.63 & \multicolumn{1}{c|}{ORT Encoder} & \multicolumn{1}{c|}{1.98} & 18.92 & \multicolumn{1}{c|}{\begin{tabular}[c]{@{}c@{}}ORT \\ Decoder\end{tabular}} & \multicolumn{1}{c|}{0.795} & 54.92 & \multicolumn{1}{c|}{120.105} & 137.47 \\ \hline
M2Transformer & \multicolumn{1}{c|}{FasterRCNN} & \multicolumn{1}{c|}{117.33} & 63.63 & \multicolumn{1}{c|}{\begin{tabular}[c]{@{}c@{}}Memory \\ Augmented \\ Encoder\end{tabular}} & \multicolumn{1}{c|}{1.093} & 10.6 & \multicolumn{1}{c|}{\begin{tabular}[c]{@{}c@{}}Meshed \\ Decoder\end{tabular}} & \multicolumn{1}{c|}{0.974} & 38.44 & \multicolumn{1}{c|}{119.397} & 112.67 \\ \hline
Transformer & \multicolumn{1}{c|}{FasterRCNN} & \multicolumn{1}{c|}{117.33} & 63.63 & \multicolumn{1}{c|}{\begin{tabular}[c]{@{}c@{}}Transformer \\ Encoder\end{tabular}} & \multicolumn{1}{c|}{1.949} & 18.92 & \multicolumn{1}{c|}{\begin{tabular}[c]{@{}c@{}}Transformer \\ Decoder\end{tabular}} & \multicolumn{1}{c|}{0.916} & 36 & \multicolumn{1}{c|}{120.195} & 118.55 \\ \hline
\end{tabular}
}
\end{adjustbox}
\end{table*}

Due to the transformer-based architectures, the computational requirements for models like \emph{ORT}~\cite{herdade2019image} and \emph{M2}~\cite{cornia2020meshed} are usually higher than other models. Real time transformer model-based image caption synthesis on resource constrained embedded devices is often challenged by high computations and large number of parameters. These high computational costs not only hinder on-device deployment but also increase reliance on cloud-based inference. This leads to latency issues and higher energy consumption. Moreover, such dependencies create accessibility barriers, particularly in remote areas where internet connectivity is unreliable. This observation motivates the present work to develop a lightweight image captioning model by utilizing low latency CNNs as visual feature encoders.

Most image captioning research has been centered around high-resource languages such as English, Chinese~\cite{ChineseFlickr8K}, and German~\cite{germenAIC}, while low-resource languages like Assamese have received minimal attention. The Assamese language is the official language of the North-East Indian state of Assam. It is an Indo-Aryan language similar to Hindi, Bengali, and Oriya. The Assamese language follows a left-to-right writing script similar to the Bengali language. Assamese is a morphologically rich language with complex grammar having inflection, gender and tense markers. Additionally, Assamese has 15.31 million native speakers, out of which ~85\% reside in rural areas and are more fluent in Assamese than English\footnote{www.language.census.gov.in/showLanguageCensusData}. A lightweight Assamese image captioning system for edge devices could address real-world challenges like assisting the visually impaired. An AI-based assistive technology using Assamese image captioning can help 0.3\% of the population of Assam who suffer from visual impairment \footnote{www.finance.assam.gov.in}. When deployed on smartphones, such a system enables on-device accessibility without relying on constant internet connectivity, making it a practical and inclusive solution for visually impaired users in resource-constrained settings. Moreover, ~70\% of students from Assam get their elementary education in the Assamese language \footnote{www.elementary.assam.gov.in/}. Therefore, a computationally efficient Assamese captioning model can help in smart education technology for language learning~\cite{hasnine2019vocabulary}. The design of image captioning models with lower computations and parameters is vital for such practical applications. This is particularly applicable for other low-resource languages as well for broader accessibility and usability of AI systems. 

% Research motivation and contribution.
While deep learning-based image captioning models have achieved remarkable success, their adoption in real-world applications remains constrained by high computational demands and hardware limitations. Most state-of-the-art models prioritize performance, often at the expense of efficiency. This makes them impractical for deployment on edge devices. This limitation is particularly evident in applications like assistive technologies for visually impaired individuals, where computational efficiency is crucial for real-time operation. To address these challenges, this work introduces  AC-Lite, a lightweight image captioning model for low-resource Assamese language. Unlike existing approaches, this is the first effort specifically aimed at developing a computationally efficient captioning model for Assamese. AC-Lite is designed for deployment on computation resource-constrained devices (e.g. smartphones). Additionally, the model should operate smoothly on the device hardware without connecting to any remote server for caption generation. To achieve this, the proposed AC-Lite model uses computationally efficient CNN models for the visual feature encoder. Moreover, the Gated Recurrent Units (GRU) have been employed in the decoder due to their fewer parameters, fewer computational requirements, and faster training times. Further, bilinear attention~\cite{mypaper_paclic} is also employed with the proposed AC-Lite to boost the performance. Finally, a reinforcement learning-based approach~\cite{selfcritical} is applied for further improvement of captioning performance. The performance of AC-Lite is evaluated by using COCO-Assamese Caption (COCO-AC) and the Flickr30K-Assamese Caption (Flickr30K-AC) datasets proposed in~\cite{mypaper_paclic}. The major contributions of this work are as follows.

\begin{itemize}

    \item Development of AC-Lite, a computationally efficient lightweight image captioning model for the Assamese language.
    
    \item Exploration of different lightweight Convolutional Neural Networks as an alternative to FasterRCNN for visual feature encoding. 
    
    \item A GRU-based caption decoder with bilinear attention is employed to achieve better performance with less computational requirement. The model performance is further enhanced using reinforcement learning. 
     
\end{itemize}

\section{Related Work}
\label{sec:formatting}

The state-of-art image captioning models adopted the encoder-decoder based deep learning framework~\cite{hossain2019comprehensive} from neural machine translation (NMT). In NMT, an RNN encodes the input word sequence, and another RNN is used as the decoder to generate the target word sequence. Vinyals \emph{et al.}~\cite{vinyals2015show} used GoogleNet~\cite{szegedy2015going} for visual feature encoding and long short-term memory (LSTM) network for caption generation. The visual features are only used as the initial hidden state for the LSTM network.  However, the visual features weaken over long sequences due to the vanishing gradient problem. Xu \emph{et al.}~\cite{xu2015show} addressed this issue through the introduction of a visual attention mechanism where attended features are fed to the LSTM at each decoding step. Anderson \emph{et al.}~\cite{anderson2018bottom} employed FasterRCNN~\cite{ren2015faster} for extracting visual features from salient regions (bottom-up features) to focus on important scene objects. Subsequent works~\cite{huang2019attention,pan2020x} reported further performance improvement using these bottom-up features. Huang \emph{et al.}~\cite{huang2019attention} incorporated multi-head self-attention to refine the visual features and combined them with LSTM for caption generation.

% Related work for transformer
\emph{Transformer architectures} have improved image captioning performance by visual feature refinement and by enhancing the semantic accuracy of captions. Cornia \emph{et al.}~\cite{cornia2020meshed} proposed a meshed-memory transformer, linking encoder and decoder layers for multi-level vision-language relationships, with a memory-augmented attention mechanism. Pan \emph{et al.}~\cite{pan2020x} introduced X-linear attention in transformers, enabling second-order interactions for richer representations. Herdade \emph{et al.}~\cite{herdade2019image} adjusted attention weights using geometric shapes and spatial positions. Nair \emph{et al.}~\cite{sid_ort} replaced trigonometric positional embeddings with spatial positions of salient regions. Notably, these transformer-based architectures are quite computation intensive (see \autoref{tab:model-computation}). 

% Lightweight Visual Feature Extractors
\emph{Lightweight Convolutional Neural Networks} have gained popularity for their operational efficiency on resource-constrained devices. For instance, MobileNetV1 ~\cite{CNN_MobileNetV1} and MobileNetV2~\cite{CNN_MobileNetV2} used techniques like depth-wise separable convolutions and inverted bottleneck blocks for reducing computations. Following that, other lightweight CNN architectures like ShuffleNetV1~\cite{CNN_ShuffleNetV1}, ShuffleNetV2~\cite{CNN_ShuffleNetV2}, EfficientNet~\cite{CNN_EfficientNet}, and MobileNetV3~\cite{CNN_MobileNetV3} were proposed. These lightweight CNNs also have their scaled-down versions like ShuffleNetV2X0.5, ShuffleNetV2X1.0, ShuffleNetV2X1.5, ShuffleNetV2X2.0 and MobileNetv3\_small which operate in different configurations. However, a trade-of between performance and computations is observed across all architectures.

% Lightweight Image Captioning
\emph{Lightweight Image Captioning} models focus on efficiency for resource constrained settings. Models like ACORT~\cite{acort} and COMIC~\cite{comic} used parameter and vocabulary size reduction techniques. ACORT~\cite{acort} reduced parameters through cross-layer and attention parameter sharing, achieving a 13$\times$ reduction compared to transformers~\cite{vaswani2017attention}. Additionally, the authors introduced Radix Encoding to minimize vocabulary size. The Radix Encoding method encodes the original vocabulary into a smaller set of symbols, thereby reducing the parameter count in the embedding matrix. COMIC~\cite{comic} employed a similar vocabulary encoding technique to reduce parameters. LCM-Captioner~\cite{LCM} used TextLighT to map visual features to a lower-dimensional space. Zhang \emph{et al.}~\cite{zhang2024mobilenet} substituted  FasterRCNN with MobileNetV3 for visual encoding and employed the transformer decoder for caption decoding. LightCap~\cite{LightCap} used pre-trained CLIP model for visual feature encoding without relying on resource-intensive object detectors. SMALLCAP~\cite{ramos2023smallcap} combined retrieval-augmented generation with pre-trained CLIP vision encoder and GPT-2, thereby reducing total trainable parameters. I-Tuning~\cite{luo2023tuning} used cross-attention to link the frozen CLIP and GPT-2 models, thus reducing trainable parameters by up to 10$\times$ while maintaining the overall performance. However, these lightweight image captioning models rely on English pre-trained language models, making them unsuitable for low-resource languages like Assamese. This motivates the need to develop lightweight captioning models for low-resource languages like Assamese. 

% Image Captioning in Indian Languages
\emph{Image Captioning in Indian Languages} like Assamese, Bengali, and Hindi is scarce, primarily due to the lack of large-scale image-caption datasets. Mishra \emph{et al.}~\cite{mishra2021hindi} introduced MSCOCO-Hindi by translating MSCOCO-English captions into Hindi. Additionally, the authors employ ResNet101~\cite{resnet101} as visual feature encoder and GRU with attention for caption generation. In another work, Mishra \emph{et al.}~\cite{mishra2021image} used the transformer decoder with ResNet101 for Hindi caption generation. In Assamese language, Nath \emph{et al.}~\cite{nath2022image} used a similar translation-based approach to create Assamese image-caption dataset and employed EfficientNetB3 with GRU for caption generation. However, automatically translated Assamese captions often suffer from syntactic and semantic errors~\cite{mypaper_paclic} and that is not addressed in~\cite{nath2022image}. Choudhury \emph{et al.}~\cite{mypaper_paclic} addressed this issue by manually correcting the translation errors to develop the \emph{COCO-Assamese} and \emph{Flickr30K-Assamese} datasets. Additionally, the authors employed FasterRCNN for visual feature encoding and Bi-LSTM with bilinear attention for caption generation. In a related study, Choudhury \emph{et al.}~\cite{mypaper_tallip} incorporated semantic attributes with visual features to improve Assamese captioning performance. The semantic attributes are key concepts presented in an image, such as object and place names, attributes, and actions. These semantic attributes are mapped to feature vectors using FastText~\cite{fasttext} embeddings trained on a large Assamese corpus. FastText was chosen for its subword-level modeling, which captures the inflectional properties of Assamese effectively. Further, Choudhury \emph{et al.}~\cite{mypaper_IJALP} demonstrated that models trained directly on Assamese datasets performed better than those trained on English datasets followed by translation into Assamese. This highlights the importance of developing image caption datasets and models for low-resource languages like Assamese.

\section{Proposed Work}
\label{sec:proposed}
\begin{figure*}
\centering
\includegraphics[width=0.7\textwidth, height=5.0cm]{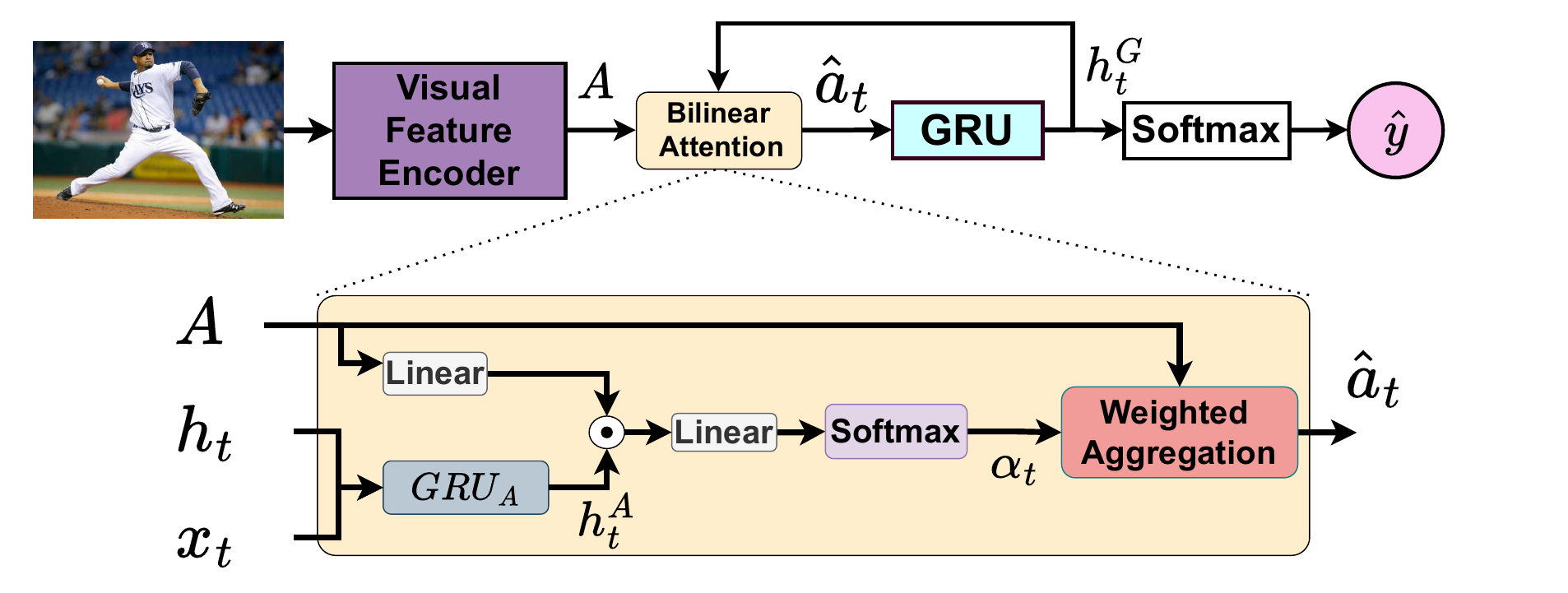}
\caption{\small{Functional block diagram of the proposed AC-Lite model with Bilinear Attention.}}
\label{fig:model-block}
\end{figure*}

The proposed AC-Lite model consists of three main components -- (a) \emph{Visual Feature Encoder}, (b) \emph{Bilinear Attention}, and (c) \emph{Gated Recurrent Unit} (GRU) for caption decoding as shown in \autoref{fig:model-block}. AC-Lite uses pre-trained CNNs (trained on ImageNet) as the visual feature encoder. Initially, the Visual Feature Encoder takes an input image and transforms it into a feature map of size $d_a \times n_h \times n_w$. These feature maps are then flattened to create visual feature vectors $\mathbf{A}=[ \mathrm{\boldsymbol{a}}_1, \ldots \mathrm{\boldsymbol{a}}_i, \ldots \mathrm{\boldsymbol{a}}_{n_{a}} ]$ ($ \mathrm{\boldsymbol{a_i}} \in \mathbb{R}^{d_a \times 1}, \mathbf{A} \in \mathbb{R}^{d_a \times n_a}, n_a = n_h \times n_w$).  The extracted visual features $\mathbf{A}$ are then fed to the Bilinear Attention module to generate the attended visual feature $\boldsymbol{\hat{a}}_t$ at each decoding step $t$.

The Bilinear Attention module has shown superior performance for Assamese image captioning task~\cite{mypaper_paclic}. In the original paper~\cite{mypaper_paclic}, the bilinear attention module consists of an attention LSTM $\mathbf{LSTM_{A}}$ and attention weight computation mechanism. However, to reduce computation, the $\mathbf{LSTM_{A}}$ in the bilinear attention module is replaced with a Gated Linear Unit (GRU) $\mathbf{GRU_{A}}$ as shown in \autoref{fig:model-block}. At the $t^{th}$ decoding step, $\mathbf{GRU_{A}}$ produces a partial caption, and its context is captured in $\mathbf{h}_{t}^{A} \in \mathbb{R}^{d_h \times 1} $. $\mathbf{GRU_{A}}$ takes mean pooled visual features $\bar{{\boldsymbol{a}}} = \frac{\sum_{i=1}^{n_a} \boldsymbol{a}_{i}}{n_a}$ concatenated with previous hidden state $\mathbf{h}_{t-1}^{G}$ of the decoder GRU as input $\mathbf{x}_{t-1} = \left [ \mathbf{h}_{t-1}^{G} : \bar{{\boldsymbol{a}}} \right ]$
to produce the partial caption.    
\begin{equation}
\label{eq:lstmatt}
\small{\mathbf{h}_{t}^{A} = \mathbf{GRU_{A}}\left ( \mathbf{x}_{t-1}, \mathbf{h}_{t-1}^{A}; \Theta^{att} \right)}
\end{equation}
\noindent Here, $\Theta^{att}$ denotes the parameters of $\mathbf{GRU_{A}}$. Subsequently, the attention computation mechanism performs low-rank bilinear pooling between visual feature $\mathbf{A}$ and $\mathbf{h}_{t}^{A}$ to generate attention weights $\boldsymbol{\alpha}_{t}=\left[\alpha_{t}[1], \dots, \alpha_{t}[i], \dots \alpha_{t}[n_a]\right]$ in the following manner. 

\begin{gather}
\label{eq:lstmatt}
\small{\beta_{t}[i]= \left\{\omega_{A}\right\}^{T}\left\{ (W_{e}^{h} \mathbf{h}_{t}^{A}) \odot (W_{e}^{a} \mathrm{\boldsymbol{a}}_i) \right\}}\\
\small{\boldsymbol{\alpha}_{t}=\mathit{SoftMax}\left\{ \left[\beta_{t}[1], \dots, \beta_{t}[i], \dots \beta_{t}[n_a]\right] \right\}}
\end{gather}

\noindent Here, $\omega_{A} \in \mathbb{R}^{d_e \times 1}$, $W_{e}^{h} \in \mathbb{R}^{d_e \times d_h}$, $W_{e}^{a} \in \mathbb{R}^{d_e \times d_a}$ are linear transformations. Later, the attended visual feature $\hat{\mathrm{\boldsymbol{a}}}_{t} \in \mathbb{R}^{d_a \times 1}$ is computed as $\small{\hat{\mathrm{\boldsymbol{a}}}_{t} = \sum_{i=1}^{n_a} \alpha_{t}[i] \mathrm{\boldsymbol{a}}_i}$.

Finally, the attended visual feature $\hat{\mathrm{\boldsymbol{a}}}_{t}$ is given as input to the language decoder GRU to produce the next word probability $\hat{\mathrm{\boldsymbol{y}}}_{t}$ conditioned on previous hidden state $\mathbf{h}_{t-1}^{G}$ as follows

\begin{gather}
\label{eq:lstmatt}
    \small{\mathbf{h}_{t}^G= \mathbf{GRU} \left( \mathrm{\boldsymbol{\hat{a}_{t}}}, \mathbf{h}_{t-1}^G; \Theta^G \right)} \\
    \small{\hat{\mathrm{\boldsymbol{y}}}_{t} = SoftMax \left( W_{o} \mathbf{h}_{t}^G \right)}
\end{gather}

\noindent Here, $W_{o} \in \mathbb{R}^{|V|\times{d_h}}$ is  a linear transformation, $\Theta^G$ are the parameters for the GRU network, and $|V|$ denotes the vocabulary size.    

\section{Experimental Setup}
\label{sec:expSetup}
The hyperparameters, dataset preparation methodology and the baseline models are briefly described in this section.

% HyperParameters
\subsection{Hyper Parameters}
\label{subsec:hp}

The visual feature encoder takes RGB images of size $3 \times 640 \times 480$ as input and produces a feature map of size $1024 \times 14 \times 14$. These features are extracted from the final convolutional layers of the CNN. Later, adaptive pooling transforms this feature map into $n_a = 14 \times 14 = 196$ feature vectors of $d_a=1024$ dimension. The GRU and attention GRU ($\mathbf{GRU_{A}}$) hidden layer sizes are set to $\mathbf{h}_{t}^{G}=\mathbf{h}_{t}^{A}= 512$. The attention embedding dimension is set to $d_e=512$. The model is trained by using Adam optimizer with a learning rate of $5 \times 10^{-4}$. The model is trained for 30 epochs with a batch size of 50 image-caption pairs.  

% Dataset Preparation
\subsection{Dataset Preparation} 
\label{subsec:dsp}

The proposed model is trained and tested with COCO-AC and Flickr30K-AC dataset. Initially, the dataset is preprocessed to remove the punctuations and the captions are truncated to a maximum of 16 tokens. Later, vocabulary lists are created from the words that occurred more than five times. This resulted in a vocabulary list of \emph{12,912} words for COCO-AC, and \emph{8,534} words for Flickr30K-AC. The COCO-AC and Flickr30K-AC dataset contains $123K$ and $31K$ images with five captions per image. The COCO-AC dataset is divided into $113K$ images for training and $5K$ each for testing and validation. For Flickr30K-AC, this division is $29K$ for training and $1K$ each for testing and validation. 

% Baseline Models
\subsection{Baseline Models} 
\label{subsec:bm}

AC-Lite is compared against the following nine baseline models.

% Show and tell
\noindent\textbf{Baseline-1}: The \emph{Baseline-1} employs the Neural Image Caption model by Vinyals \emph{et al.} \cite{vinyals2015show}. This model uses ResNet101 as the visual feature encoder and LSTM as the caption decoder with no attention module. 

%Show attend and tell
\noindent\textbf{Baseline-2}: The  \emph{Baseline-2} employs the Show Attend and Tell model by Xu \emph{et al.} \cite{xu2015show}. This model uses soft attention with LSTM as the caption decoder and ResNet101 as the visual feature encoder. 

%BUTD
\noindent\textbf{Baseline-3}: The  \emph{Baseline-3}  employs the Bottom Up and Top Down model by Anderson \emph{et al.} \cite{anderson2018bottom}. This model uses a bottom-up attention module with the LSTM caption decoder and a pre-trained FasterRCNN as the visual feature encoder.

%AOANET
\noindent\textbf{Baseline-4}: The  \emph{Baseline-4} employs the Attention on Attention Network (AoAN-et) by Huang \emph{et al.}~\cite{huang2019attention}. This model uses the Attention on Attention (AoA) module, which consists of a multi-head self-attention followed by an additional attention layer. The AoA module is used for visual feature refinement and with the LSTM decoder for caption generation.  

%ORT
\noindent\textbf{Baseline-5}: The  \emph{Baseline-5} employs the Object Relation Transform (ORT) model by Herdade \emph{et al.}~\cite{herdade2019image}. This model uses the geometric position of objects in the transformer encoder instead of positional encoding. These geometric features adjust attention weights and improve the ability of the model to capture spatial relationships within images for more accurate caption generation.

%M2
\noindent\textbf{Baseline-6}: The \emph{Baseline-6} uses the Mesh-Memory Transformer by Cornia \emph{et al.}~\cite{cornia2020meshed}, which connects encoder and decoder sub-layers in a mesh structure to capture multi-level visual relationships. It serves to evaluate whether integrating semantic information offers improvements over memory-based connections.

%Basic Transformers
\noindent\textbf{Baseline-7}: The \emph{Baseline-7} employs the standard transformer model introduced by Vaswani \emph{et al.}~\cite{vaswani2017attention}, selected for its foundational role in sequence generation. It uses an encoder to refine visual features, which are then decoded into captions. Six layers of encoders and decoders are stacked to facilitate fair comparison with models incorporating semantic and spatial enhancements.

%Bi-LSTM (PACLIC)
\noindent\textbf{Baseline-8}: The  \emph{Baseline-8} employs a Bi-LSTM network with Bilinear Attention~\cite{mypaper_paclic} for Assamese image caption generation. The Bi-LSTM captures sequential dependencies in both forward and backward directions which is important for capturing the linguistic properties of Assamese language. Additionally, the bilinear attention applies second-order interactions to improve model performance. 

%VSSA (TALLIP)
\noindent\textbf{Baseline-9}: The  \emph{Baseline-9} integrates semantic attributes with visual features~\cite{mypaper_tallip} using VSSA module for Assamese image caption generation. The VSSA module uses separate self-attention submodule to process semantic attributes with visual features. Additionally, a Bi-GRU is employed to capture both forward and backward contextual information, which is particularly important for modeling the linguistic characteristics of the Assamese language.

% Results and Discussion
\section{Results and Discussion}
\label{sec:expRes}
The quantitative performance analysis, ablation studies and qualitative results are presented in this section.

% Quantitative Results
\subsection{Quantitative Results}
\label{subsec:qnt}

The quantitative results are obtained by using beam sizes of 6 and 3 for COCO-AC and Flickr30K-AC, respectively. The \autoref{tab:coco-ac} and \autoref{tab:f30K-ac} compare the performance of the proposed AC-Lite model with baseline models. Note that the computation-complexity and number of parameters are different for the same model in \autoref{tab:coco-ac} and \autoref{tab:f30K-ac} because of the different vocabulary sizes for COCO-AC and Flickr30K-AC. 

\begin{table*}[]
\centering
\caption{Performance of AC-Lite model compared against nine baselines. All models are trained and tested on the COCO-AC dataset. AC-Lite+XE and AC-Lite+RL respectively denote cross-entropy loss minimization and reinforcement learning based model training. Performances are reported in terms of BLEU-n, CIDEr scores, computations (GFLOPs) and parameters in millions (M).}
\label{tab:coco-ac}
\begin{adjustbox}{width=\textwidth}{
\begin{tabular}{|c|c|c|c|c|c|c|c|}
\hline
\textbf{Models} & \textbf{BLEU-1} & \textbf{BLEU-2} & \textbf{BLEU-3} & \textbf{BLEU-4} & \textbf{CIDEr} & \textbf{GFLOPs} & \textbf{Params (M)} \\ \hline
Baseline-1 & 59.9 & 41.5 & 29.1 & 20.6 & 61.5 & 48.121 & 59.41 \\ \hline
Baseline-2 & 62.1 & 45.1 & 31.8 & 23.2 & 68.3 & 48.392 & 60.46 \\ \hline
Baseline-3 & 67.1 & 49.6 & 36.5 & 27 & 80.8 & 118.336 & 122.61 \\ \hline
Baseline-4 & 67.3 & 50.1 & 37 & 27.6 & 81.5 & 122.839 & 156.21 \\ \hline
Baseline-5 & 66.3 & 48.8 & 35.7 & 26.2 & 79.7 & 120.105 & 137.47 \\ \hline
Baseline-6 & 66.9 & 49.5 & 36.6 & 27.4 & 80.4 & 119.398 & 112.67 \\ \hline
Baseline-7 & 67.4 & 49.6 & 36.3 & 26.8 & 81.1 & 120.227 & 122.06 \\ \hline
Baseline-8 & 68.1 & 50.5 & 37.2 & 27.5 & 81.7 & 119.578 & 173.72 \\ \hline
Baseline-9 & 68.6 & 51 & 38.1 & 28.6 & 84.5 & 120.01 & 178.04 \\ \hline
AC-Lite + XE & 63.6 & 45.8 & 33.1 & 24.1 & 72 & 2.45 & 22.87 \\ \hline
AC-Lite + RL & \textbf{68.3} & \textbf{50.2} & \textbf{36.2} & \textbf{26} & \textbf{82.3} & 2.45 & 22.87 \\ \hline
\end{tabular}
}
\end{adjustbox}
\end{table*}

\begin{table*}[]
\centering
\caption{Performance of AC-Lite model compared against nine baselines. All models are trained and tested on the Flickr30K-AC dataset. AC-Lite+XE and AC-Lite+RL respectively denote cross-entropy loss minimization and reinforcement learning based model training. Performances are reported in terms of BLEU-n, CIDEr scores, computations (GFLOPs) and parameters in millions (M).}
\label{tab:f30K-ac}
\begin{adjustbox}{width=\textwidth}{
\begin{tabular}{|c|c|c|c|c|c|c|c|}
\hline
\textbf{Models} & \textbf{BLEU-1} & \textbf{BLEU-2} & \textbf{BLEU-3} & \textbf{BLEU-4} & \textbf{CIDEr} & \textbf{GFLOPs} & \textbf{Params (M)} \\ \hline
Baseline-1 & 51.1 & 32.5 & 21 & 13.3 & 31.9 & 48.081 & 54.92 \\ \hline
Baseline-2 & 53.3 & 35.6 & 24.1 & 16.5 & 39.7 & 48.352 & 55.98 \\ \hline
Baseline-3 & 60.4 & 42.3 & 29.5 & 20.1 & 46.5 & 118.258 & 113.85 \\ \hline
Baseline-4 & 61 & 42.7 & 30.1 & 21.2 & 48.6 & 122.825 & 147.24 \\ \hline
Baseline-5 & 60.1 & 41.8 & 28.7 & 20 & 47.2 & 120.013 & 115.47 \\ \hline
Baseline-6 & 59.5 & 42.1 & 29.4 & 20.1 & 47.4 & 119.332 & 100.36 \\ \hline
Baseline-7 & 60.6 & 43 & 30.3 & 20.9 & 48 & 120.186 & 117.58 \\ \hline
Baseline-8 & 61.5 & 43.4 & 30.4 & 21.1 & 48.3 & 119.42 & 164.2 \\ \hline
Baseline-9 & 62.7 & 45.2 & 32.3 & 23 & 51.4 & 119.937 & 167.77 \\ \hline
AC-Lite + XE & 54.9 & 37 & 25.1 & 16.6 & 40 & 2.41 & 18.38 \\ \hline
AC-Lite + RL & \textbf{58.5} & \textbf{40.2} & \textbf{27.9} & \textbf{18.4} & \textbf{46.1} & 2.41 & 18.38 \\ \hline
\end{tabular}
}
\end{adjustbox}
\end{table*}

As illustrated in \autoref{tab:coco-ac}, the AC-Lite model trained with cross-entropy loss (AC-Lite + XE) achieves respective BLEU-1, BLEU-4, and CIDEr scores of 63.6, 24.1, and 72 on COCO-AC with only 2.45 GFLOPs and 22.87M parameters. For Flickr30K-AC, the AC-Lite model achieves 54.9 BLEU-1, 16.6 BLEU-4, and 40.0 CIDEr score with only 2.41 GFLOPs and 18.38M parameters. Compared to specifically designed Assamese image captioning models Baseline-8 and Baseline-9, the performance of the AC-Lite model is slightly lower. However, the computation requirement is approximately 48$\times$ lower than these Assamese captioning models. Moreover, when the proposed AC-Lite is trained with reinforcement learning (AC-Lite + RL) achieving 68.3 BLEU-1, 26 BLEU-4, and 82.3 CIDEr on COCO-AC. In the case of Flickr30K-AC, the model trained with reinforcement learning (AC-Lite + RL) gives competitive results achieving 58.5 BLEU-1, 18.4 BLEU-4, and 46.1  CIDEr. This highlights that AC-Lite achieves a favorable trade-off between performance and efficiency, making it suitable for deployment in resource-constrained applications.

\begin{table*}[]
\centering
\caption{Performance of the proposed model with different pretrained CNN backbone based visual feature encoders. GRUs are used for both attention and decoder units. All models are trained and tested on the COCO-AC dataset by cross-entropy loss minimization. Performances are reported in terms of BLEU-n, CIDEr scores, computations (GFLOPs) and parameters in millions (M).}
\label{tab:best-cnn}
\begin{adjustbox}{width=\textwidth}{
\begin{tabular}{|c|c|c|c|c|c|c|c|}
\hline
\textbf{Visual Feature Encoder} & \textbf{BLEU-1} & \textbf{BLEU-2} & \textbf{BLEU-3} & \textbf{BLEU-4} & \textbf{CIDEr} & \textbf{GFLOPs} & \textbf{Params (M)} \\ \hline
ShuffleNetv2x0.5 & 57.7 & 39.6 & 27.6 & 19.3 & 59.1 & 0.865 & 20.730 \\ \hline
MobileNetV3\_Small & 56.6 & 38.5 & 26.1 & 17.8 & 55.6 & 0.964 & 21.32 \\ \hline
ShuffleNetv2x1.0 & 61.1 & 43.3 & 30.8 & 22 & 66.2 & 1.516 & 21.64 \\ \hline
MobileNetV3\_Large & 59.5 & 41.6 & 28.7 & 19.9 & 61.6 & 1.972 & 23.36 \\ \hline
ShuffleNetv2x1.5 & \textbf{63.6} & \textbf{45.8} & \textbf{33.1} & \textbf{24.1} & \textbf{72} & 2.45 & 22.87 \\ \hline
MobileNetV2 & 56.5 & 38.6 & 26.8 & 18.8 & 55.7 & 2.527 & 22.610 \\ \hline
EfficientNetB0 & 59.6 & 42 & 29.1 & 20.1 & 61.6 & 3.059 & 24.4 \\ \hline
ShuffleNetv2x2.0 & 62.8 & 45.2 & 32.5 & 23.4 & 70.5 & 4.216 & 25.73 \\ \hline
EfficientNetB1 & 59.6 & 41.9 & 29 & 20.2 & 63.1 & 4.221 & 26.9 \\ \hline
ResNet18 & 63.4 & 45.7 & 32.9 & 23.8 & 72.2 & 11.745 & 31.57 \\ \hline
ResNet50 & 65.7 & 48.1 & 35 & 25.6 & 76.8 & 25.771 & 43.9 \\ \hline
ResNet101 & 66.4 & 48.8 & 35.7 & 26.1 & 78.5 & 48.562 & 62.89 \\ \hline
\end{tabular}
}
\end{adjustbox}
\end{table*}

% Ablation Analysis
\subsection{Ablation Analysis}
\label{subsec:abls}

The \autoref{tab:best-cnn} presents an ablation study evaluating different CNN architectures as visual feature encoders for the proposed model. The experiments aim to identify the encoder that provides the best performance while maintaining low computational requirements. Among lightweight CNN architectures, ShuffleNetv2x1.5 achieves the best performance, with BLEU-1, BLEU-4, and CIDEr scores of 63.6, 24.1, and 72.0, respectively, at total computation costs of 2.45 GFLOPs and 22.87M parameters. ShuffleNetv2x1.0 and ShuffleNetv2x2.0 also perform well, with CIDEr scores of 66.2 and 70.5, respectively, demonstrating the efficiency of the ShuffleNetv2 family. EfficientNetB0 and EfficientNetB1 yield moderate results but require higher FLOPs and parameters compared to ShuffleNet variants. ResNet101 achieves the highest performance, with a CIDEr score of 78.5, but its computational cost is significantly higher, requiring 48.581 GFLOPs and 63.94 million parameters. This analysis indicates that ShuffleNetv2x1.5 provides an effective trade-off between performance and computational efficiency, making it a viable encoder option for lightweight image captioning models.

\begin{table*}[]
\centering
\caption{Performance of the proposed model with different RNNs for attention and language decoding. ShuffleNetV2x1.5 is used as the visual feature encoder. The model is trained and tested on the COCO-AC dataset by cross-entropy loss minimization. Performances are reported in terms of BLEU-n, CIDEr scores, computations (GFLOPs) and parameters in millions (M).}
\label{tab:ablation_rnn}
\begin{adjustbox}{width=\textwidth}{
\begin{tabular}{|c|c|c|c|c|c|c|}
\hline
\textbf{Attention RNN} & \textbf{Language RNN} & \textbf{BLEU-1} & \textbf{BLEU-4} & \textbf{CIDEr} & \textbf{GFLOPs} & \textbf{Params (M)} \\ \hline
LSTM & LSTM & 62.8 & 23.9 & 71 & 2.483 & 24.71 \\ \hline
LSTM & GRU & 63 & 23.5 & 69.8 & 2.469 & 23.92 \\ \hline
GRU & LSTM & 62.7 & 23.4 & 70 & 2.464 & 23.65 \\ \hline
GRU & GRU & \textbf{63.6} & \textbf{24.1} & \textbf{72} & 2.45 & 22.87 \\ \hline
\end{tabular}
}
\end{adjustbox}
\end{table*}

\begin{figure*}
\centering
\includegraphics[width=\textwidth, height=7.5cm]{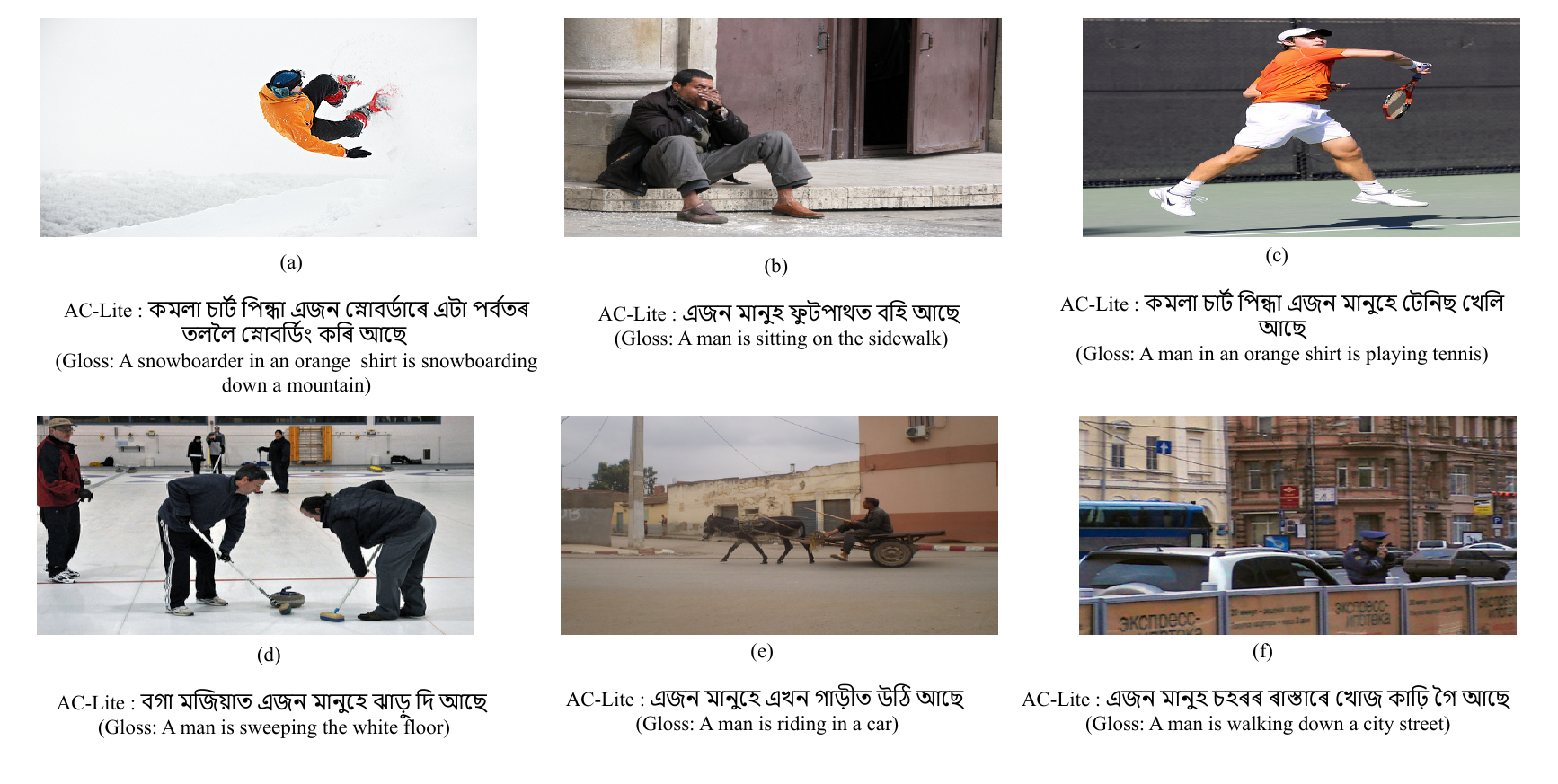}
\caption{\small{Qualitative example produced by the proposed AC-Lite on COCO-AC test set. Here, AC-Lite -- caption generated by the proposed model, gloss -- gloss annotation.}}
\label{fig:quality}
\end{figure*}

Another ablation study was conducted to assess the impact of different types of RNNs used in the attention and language modules of the proposed model. For this ablation study all experiments were performed using ShuffleNetV2x1.5 as the visual feature encoder and evaluated on the COCO-AC dataset.  As shown in Table~\ref{tab:ablation_rnn}, the combination of GRU for both the attention and language modules yields the best performance, achieving the highest BLEU-1 as 63.6, BLEU-4 as 24.1, and CIDEr score 72, while also maintaining the lowest computational cost at 2.45 GFLOPs and 22.87 million parameters. In comparison, replacing either GRU with LSTM leads to drops in performance and increased computational load. Notably, the LSTM-LSTM configuration, while achieving a slightly lower CIDEr score of 71, is the most resource-intensive with 2.483 GFLOPs and 24.71M parameters. These results indicate that the GRU-GRU combination improves efficiency and enhances the overall captioning quality, making them the preferred choice for lightweight and effective image captioning models.

% Qualitative Analysis
\subsection{Qualitative Analysis}
\label{subsec:qla}

The \autoref{fig:quality} shows qualitative examples of captions generated by the proposed AC-Lite model for the test set of the COCO-AC dataset. For example, as illustrated in \autoref{fig:quality}(a) and \autoref{fig:quality}(b), the AC-Lite model can identify the color of the shirt as 
% ``\bnword{কমলা}'' (Orange) as well as activity ``\bnword{স্নোবৰ্ডিং} \hspace{1em}'' (Snowboarding) and ``\bnword{টেনিছ খেলি আছে}\hspace{1.5em}'' (Playing Tennis). Additionally, in \autoref{fig:quality}(a), the proposed model can identify the environment as ``\bnword{পৰ্বত}'' (Mountain) more accurately. The same can be observed with \autoref{fig:quality}(b) and \autoref{fig:quality}(f) where the model accurately identifies the surrounding environment ``\bnword{ফুটপাথ}'' (Sidewalk) and ``\bnword{চহৰ}'' (City) respectively. 
\includegraphics[width=1cm, height=.4cm]{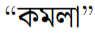}
% ``{\bengalifont কমলা}'' 
(Orange) as well as activity 
\includegraphics[width=1.2cm, height=.4cm]{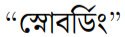}
% ``{\bengalifont স্নোবৰ্ডিং}'' 
(Snowboarding) and 
\includegraphics[width=2.2cm, height=.4cm]{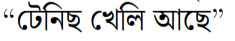}
% ``{\bengalifont টেনিছ খেলি আছে}'' 
(Playing Tennis). Additionally, in \autoref{fig:quality}(a), the proposed model can identify the environment as 
\includegraphics[width=1cm, height=.4cm]{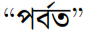}
% ``{\bengalifont পৰ্বত}'' 
(Mountain) more accurately. The same can be observed with  \autoref{fig:quality}(b) and  \autoref{fig:quality}(f) where the model accurately identifies the surrounding environment 
\includegraphics[width=1.2cm, height=.4cm]{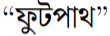}
% ``{\bengalifont ফুটপাথ}''
(Sidewalk) and 
\includegraphics[width=1cm, height=.4cm]{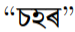}
% ``{\bengalifont চহৰ}'' 
(City) respectively.
This is attributed to the fact that the CNN models provide the whole image as input to the language decoder instead of a few selected image regions. However, inputting all image regions has drawbacks, such as the proposed model failing in object counting and identification. For instance, in \autoref{fig:quality}(d), the proposed model counts one person. Similarly, the proposed model fails to identify the ``horse'' and ``car'' present in the \autoref{fig:quality}(e) and \autoref{fig:quality}(f), respectively.

% Conclusion
\section{Conclusion}
\label{sec:conc}

This paper introduces AC-Lite, a lightweight image captioning model designed for the low-resource Assamese language. The AC-Lite model is designed to meet the need for computationally efficient solutions that can operate on resource-constrained devices. AC-Lite employs lightweight ShuffleNetv2x1.5 for visual feature extraction and Gated Recurrent Units (GRUs) in the decoder. This incorporation significantly reduces computational demands and model parameters. Additionally, the integration of bilinear attention further enhances captioning performance. AC-Lite is designed to function independently on device hardware, thereby eliminating dependency on remote servers. This makes it suitable for real-world applications like assistive technologies and smart education technology for Assamese speakers, particularly in remote areas having issues with continuous high-speed internet connectivity. 

Future efforts will primarily focus on the following important directions. First, the proposal of novel lightweight image feature extractor and language decoder architectures along with efficient training strategies. Second, the development of application-specific image-caption datasets for assistive healthcare and education technologies in Indian contexts. Finally, system-level implementation of the lightweight image captioning model on resource-constrained edge devices for application specific deployment.

{\small
\bibliographystyle{ieee}
\bibliography{AIC_PC_YA_IJCNN_2024}
}

\end{document}